\documentclass[lettersize,journal]{IEEEtran}
\usepackage{amsmath,amsfonts}
\usepackage{algorithmic}
\usepackage{algorithm}
\usepackage{array}
\usepackage[caption=false,font=normalsize,labelfont=sf,textfont=sf]{subfig}
\usepackage{textcomp}
\usepackage{stfloats}
\usepackage{url}
\usepackage{verbatim}
\usepackage{graphicx}
\usepackage{cite}
\usepackage{float}

\usepackage{cite}
\usepackage{amsmath,amssymb,amsfonts}
\usepackage{xcolor}
\usepackage{bm}
\begin{document}

\title{Deep Photonic Reservoir Computer for Speech Recognition}

\author{\IEEEauthorblockN{Enrico Picco\IEEEauthorrefmark{1},
Alessandro Lupo\IEEEauthorrefmark{1}, Serge Massar\IEEEauthorrefmark{1}}

\IEEEauthorblockA{\IEEEauthorrefmark{1}Laboratoire d’Information Quantique, CP 224, Université Libre de Bruxelles (ULB), B-1050, Bruxelles, Belgium}
}



\maketitle
\begin{abstract}
Speech recognition is a critical task in the field of artificial intelligence and has witnessed remarkable advancements thanks to large and complex neural networks, whose training process typically requires massive amounts of labeled data and computationally intensive operations. 
An alternative paradigm, reservoir computing, is energy efficient and is well adapted to implementation in physical substrates, but exhibits limitations in performance when compared to more resource-intensive machine learning algorithms. In this work we address this challenge by investigating different architectures of interconnected reservoirs, all falling under the umbrella of deep reservoir computing. We propose a photonic-based deep reservoir computer and evaluate its effectiveness on different speech recognition tasks. We show  specific design choices that aim to simplify the practical implementation of a reservoir computer while simultaneously achieving high-speed processing of high-dimensional audio signals. Overall, with the present work we hope to help the advancement of low-power and high-performance neuromorphic hardware.
\end{abstract}



\section{Introduction}

Artificial Intelligence (AI) has deeply transformed the way we think about computing. The disruptive technological revolution initiated by AI is permeating broad sectors of our everyday life: AI algorithms have been successfully applied to various domains, such as natural language processing, cyber security, financial forecasting and healthcare. 

While the results of Artificial Neural Networks (ANNs) in many applications are unmatched to other computing approaches, these networks still require huge amount of energy to be trained \cite{reuther2020survey} with respect to their biological counterpart, i.e. the human brain \cite{mehonic2022brain}. Even though ANNs can indeed mimic the high-level functionality of the brain, there is still a long way to the creation of power-efficient biological-inspired networks. 

Reservoir Computing (RC) \cite{jaeger2004harnessing, maass2002real} is a subset of ANNs, well suited for applications involving time series such as speech recognition, video recognition, time series forecasting, see \cite{tanaka2019recent} for a review
A reservoir computer is a recurrent ANN where interconnections between neurons are fixed (e.g.\ selected at random) and only a linear output layer is trained. This approach both allows to avoid the power-expensive procedure of back-propagation during training \cite{lukovsevivcius2009reservoir} and facilitates the implementation of the algorithm on unconventional computing substrates \cite{tanaka2019recent}. 
]RC has been used in a broad range of applications, such as time series forecasting \cite{jaeger2002adaptive},  emulation of chaotic systems \cite{antonik2018using},  compensation of the distortion in a nonlinear communication channel \cite{jaeger2004harnessing}, image classification \cite{jalalvand2015real} and pattern classification using different types of data such as waveforms \cite{paquot2012optoelectronic}, audio samples \cite{verstraeten2005isolated} and video sequences \cite{picco2023high}. 
However, in terms of performance, RC still lags behind its power-hungry competitors such as deep and fully-trained ANNs.

In order to bridge this disparity in performance, several approaches have been investigated. One possibility is to act on the training procedure. For example by reintroducing back-propagation, as demonstrated in \cite{hermans2014optoelectronic} and experimentally shown in \cite{hermans2015photonic, hermans2015trainable, hermans2016embodiment }, or by including knowledge about the dynamics of computational substrate in the training procedure, as in the ``physics-aware'' training scheme proposed in \cite{wright2022deep}.
A second possibility, which is the one pursued in this work, is to concatenate multiple reservoirs with the purpose of forming a more powerful network. In particular, it is claimed that connecting several reservoirs in series delivers better performance than simply enlarging a single reservoir. The approach, known as Deep Reservoir Computing (DRC), has been first presented in \cite{triefenbach2010phoneme, freiberger2019improving} and discussed more formally in  \cite{gallicchio2017deep, ma2020deepr}. Previous works have highlighted how DRC outshines standard RC in computation capabilities \cite{gallicchio2018design, gallicchio2016deep, gallicchio2018short} thanks to enhanced memory capacity, increased richness in the reservoir dynamics, and multiple time-scale representation of the input signal.
Attention on hierarchical networks of reservoirs is growing: in \cite{Na2023}, authors propose a sparse learning strategy to address deep reservoir networks and solve multidimensional chaotic time series prediction; Pedrelli et al. \cite{Pedrelli2022} propose a novel architecture based on hiearchical reservoirs to tackle real-time applications, and test it on speech recognition; in another recent work \cite{Chang2022}, the authors propose a deep reservoir architecture that can adapt to fast changing environments with limited training data.
However, these new approaches are usually investigated with software implemented on silicon chips, while there are still few implementations of these algorithms on physical substrates. The present work aims to be a stepping stone in bridging the gap between numerical and experimental implementation of deep reservoir computing, within the field of photonic neuromorphic computing.

Implementations of ANNs in photonic systems are getting an increasing interest due to several advantages, such as high-bandwidth \cite{xu202111,feldmann2021parallel}, parallelization capabilities \cite{liutkus2014imaging, saade2016random} and immunity to electromagnetic interference \cite{lugnan2020photonic}.
The first photonic experimental demonstration of DRC has been reported by Nakajima et al.\ in \cite{nakajima2022physical}. The authors concatenate multiple reservoirs and focus their attentions on the training of inter-reservoir connections. 
In a more recent work \cite{lupo2023fully}, we presented a photonic DRC based on a  wavelength-multiplexing system. 
In the present work we pursue the study of DRC by employing an optoelectronic setup based on the well-known time-multiplexing scheme\cite{appeltant2011information, larger2012photonic, paquot2012optoelectronic}. Despite its simplicity, time-delay reservoir continues to gather much attention even in the very recent past \cite{Vettelschoss2022, Köster2022, Tang2023, Estebanez2023 }, and the first implementations on integrated optical chips begin to emerge \cite{Abdalla2023}.

In this work, we pursue the experimental investigation of how DRC architectures can enhance the computing performance by (i) stacking multiple reservoir layers in series, and (ii) optimizing the interconnections between the layers. 
We test our system on two tasks involving the processing of human speech: the recognition of spoken digits, and the recognition of speakers. 
We also implement a recent design technique to improve the robustness and flexibility of delay RC systems introduced through numerical simulations in \cite{hulser2022role}. We show that with this system it is possible to classify human speech in real-time, at the same time retaining the computational benefits of DRC.

The paper is structured as follows. Sec.\ \ref{sec_RC} introduces the basic principles of RC, DRC and delay-based RC. 
The experimental setup, optimization algorithms, design choices and datasets are explained in Sec.\ \ref{sec_exp}.
Sec.\ \ref{sec_results} contains the presentation and discussion of our experimental results, and Sec.\ \ref{sec_conclusion} concludes the paper. 

\section{Reservoir Computing}
\label{sec_RC}

\subsection{``Shallow'' Reservoir Computing}

\begin{figure}[b]
\centerline{\includegraphics[scale=0.5]{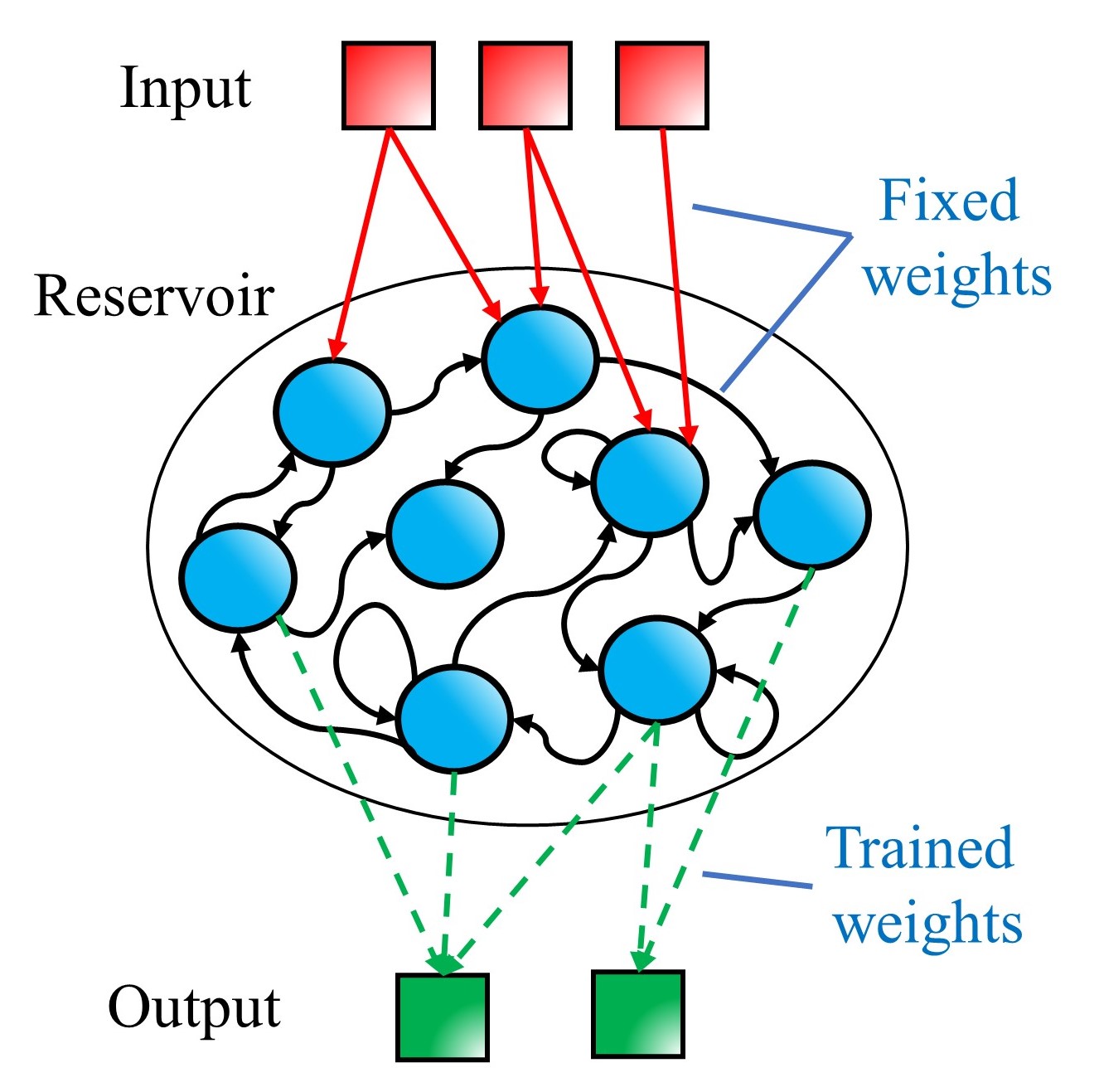}}
\caption{Architecture of a standard, or ``shallow'', reservoir computer. The fixed input and reservoir connections are represented with solid arrows, whereas the trained output connections with dashed arrows.}
\label{fig_RC}
\end{figure}

Reservoir Computing (RC) is a machine learning paradigm that falls under the umbrella of Recurrent Neural Networks (RNNs), which are designed to process time series. 
The key difference between RC and traditional RNNs is that a RC is mainly constituted of a set of neurons, known as ``reservoir'', whose interconnections are selected at random and not trained. A trained (often linear) output layer is connected to the reservoir. 
In this scheme, the reservoir behaves as a high-dimensional nonlinear feature extractor that maps the time dependent input signal to a higher dimensional space, where the subsequent (linear) output layer can construct the desired output signal. The basic architecture of a Reservoir Computer is shown in Fig.\ \ref{fig_RC}.

A reservoir composed of $N$ neurons is represented by a $N\times N$ interconnection matrix $\mathbf{W}$. The neurons in the reservoir usually exhibit a nonlinear activation function $f$, such as the hyperbolic tangent or sigmoid function. We assume the input data to be encoded in a time-dependent vector of size $K$, $\mathbf{u}(n)$, where $n$ represents the discrete time ($n\in \mathbb{Z}$). The input-to-reservoir connectivity is represented by a $N\times K$ random matrix $\mathbf{W}_\textit{in}$, usually referred to as "input mask". The values of $\mathbf{W}$ and $\mathbf{W}_\textit{in}$ are usually drawn from a uniform distribution between $[-1, +1]$.
We define $\mathbf{x}(n)$ as the time-dependent $N$-dimensional vector of neuronal signals from the reservoir, where $n$ again represents the discrete time. The evolution of the reservoir signals $\mathbf{x}(n)$ is then described as:
\begin{equation} 
\label{eq_RC}
\mathbf{x}(n+1) = f\left( \mathbf{W}\cdot \mathbf{x}(n) + \mathbf{W}_\textit{in}\cdot\mathbf{u}(n+1)\right).
\end{equation}
After the reservoir computer is run with the training data, all the state vectors are collected in the $N$x$K$ state matrix $\mathbf{X}$. Knowing the reservoir evolution $\mathbf{X}$ it is possible to estimate the optimal output weights $\mathbf{w}$ such that the output reproduces a certain desired signal $\mathbf{\tilde y}$ (known as ``target'' signal). In this work we use the regularized linear regression \cite{tikhonov1995numerical}:
\begin{equation}
w = (\mathbf{X}^T\mathbf{X}+\lambda\mathbf{I})^{-1}\mathbf{X}^T \mathbf{\tilde y},
\label{eq_ridge}
\end{equation}
where $T$ represent the transposition operation and $\lambda$ is the regularization parameter. In the case of a classification with $C$ output classes, $\mathbf{w}$ is a $C\times N$ matrix.
The output $\mathbf{y}$ of the reservoir is then computed by
\begin{equation} \label{eq_y}
\mathbf{y}(n) = \sum_{i=0}^{N-1} w_i x_i(n)
\end{equation}
and used to evaluate the performance of the reservoir computer, choosing the appropriate figure of merit for the task under test.
Some design parameters, such as the spectral radius and the connectivity of $\mathbf{W}$ or the norm of $\mathbf{W}_\textit{in}$, have a great impact on the reservoir's performance: they are referred to as hyperparameters of the reservoir. More detailed discussion about the hyperparameters and their selection can be found in Sec.\ \ref{sec_delay} and Sec.\ \ref{sec_hp}.

\subsection{``Deep'' Reservoir Computing}

Deep Reservoir Computing (DRC) is a recent extension of traditional RC in which multiple reservoir layers are stacked on top of each other\cite{gallicchio2017deep}.

The basic principle of a DRC with $L$ layers is described as follows. The first reservoir is driven by the masked input signal, following Eq.\ \eqref{eq_RC}. The states of the reservoir are mapped through a $N$x$N$ interlayer mask $\mathbf{W}_l$ and used to drive the second reservoir. The states of the second reservoir are again mapped through $\mathbf{W}_l$ and fed to the third reservoir. This process is repeated until the $L$-th reservoir, and then the states of all the reservoir layers are used for the training. 
The evolution in discrete time $n\in \mathbb{Z}$ of this deep reservoir computer follows Eq.\ \eqref{eq_RC} for the first layer, whereas for layers $l=2,...,L$  can be described as:
\begin{equation} \label{eq_DRC}
\mathbf{x}^{l}(n+1) = f\left(\mathbf{W} \mathbf{x}^l(n) + \mathbf{W}_l \mathbf{x}^{l-1}(n+1)\right)
\end{equation}
where $\mathbf{x}^{l}(n)$ is the $N$-sized state vector of the $l$-th layer. 
The values of the $L$ interlayer weights matrixes $W_l$ are randomly and independently drawn from a uniform distribution between $[-1, +1]$. Thus, the  $W_l$  are not equal. 
The training principle is the same as a ``shallow'' reservoir (Eq.\ \eqref{eq_ridge}), but the state matrix $\mathbf{X}$ is obtained by concatenating the state matrixes of all the $L$ layers, of size $N\times K$ each:
\begin{equation} \label{eq_X}
\mathbf{X} = \left[\mathbf{X}^{1},\, \mathbf{X}^{2},\, ...,\,  \mathbf{X}^{l},\, ...,\,  \mathbf{X}^{L}\right],
\end{equation}
meaning that the size of $\mathbf{X}$ is now $L\times N\times K$. 
It follows that also the size of $\mathbf{w}$ increases from $N$ to $L\cdot N$. The weights are then used to compute the output in the same way as the standard reservoir, following Eq.\ \eqref{eq_y}. 

\begin{figure}[!t]
\centerline{\includegraphics[scale=0.4]{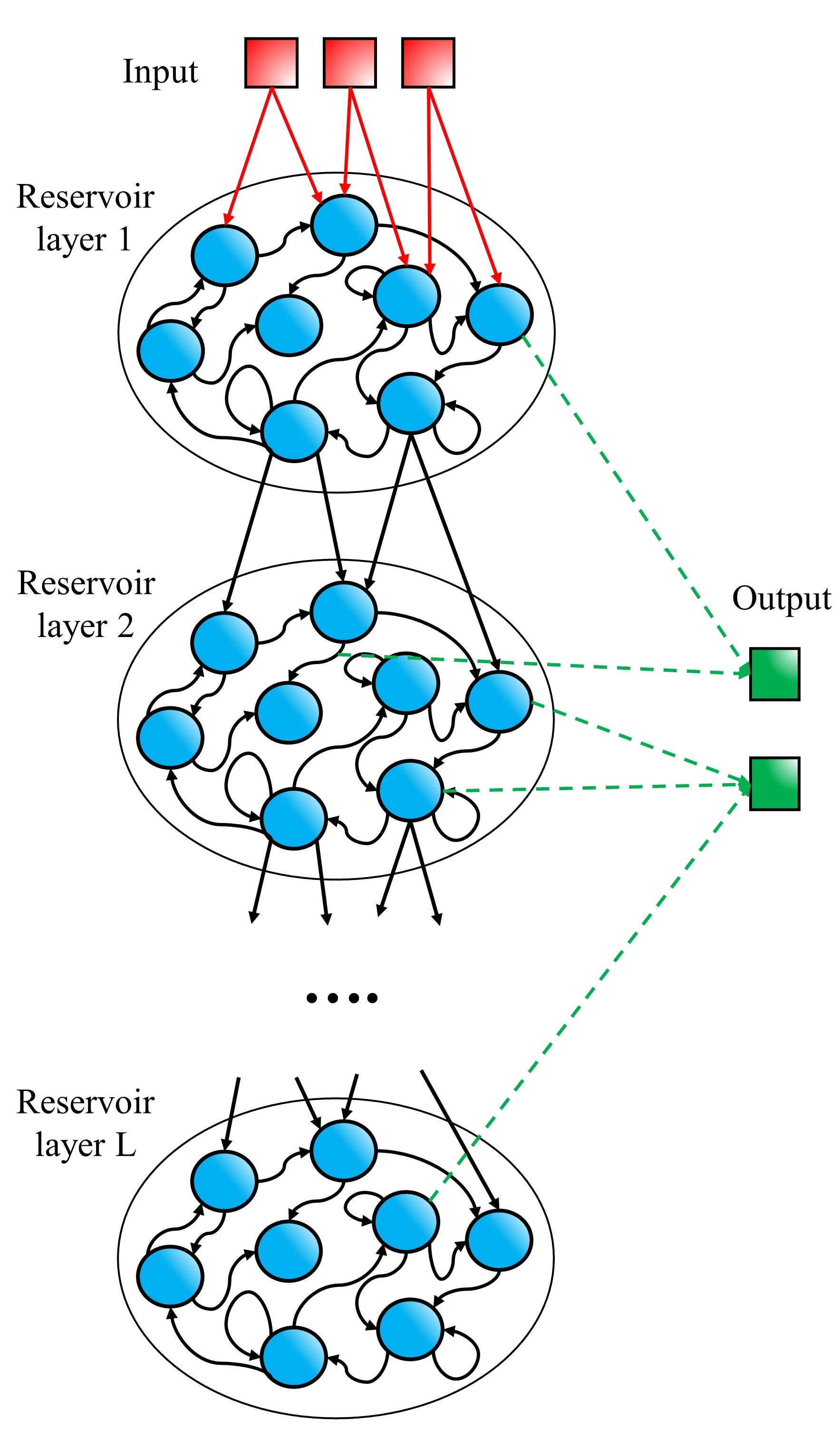}}
\caption{Architecture of a deep reservoir computer. Solid arrows represent fixed interconnections; dashed arrows represent trained interconnections. The grey solid arrows between the reservoir layers represent the random untrained interlayer connections, described by matrix $\mathbf{W}_l$.}
\label{fig_DRC}
\end{figure}

In this work we study two different DRC configurations. 
The first one is shown in Fig.\ \ref{fig_DRC} and consists of $L$ layers connected by an interlayer mask $\mathbf{W}_l$ with fixed random values, and it follows the principles described up to now. 
The second DRC configuration is shown in Fig.\ \ref{fig_DRC_EA} and contains only two layers. The main difference is that the two layers are connected by an interlayer mask whose coefficients are not fixed and random, but are optimized by means of an Evolutionary Algorithm (EA). More details about how the EA is implemented are reported in Sec.\ \ref{sec_EA}.

\subsection{Delay-based RC}
\label{sec_delay}

\begin{figure}[!b]
\centerline{\includegraphics[scale=0.5]{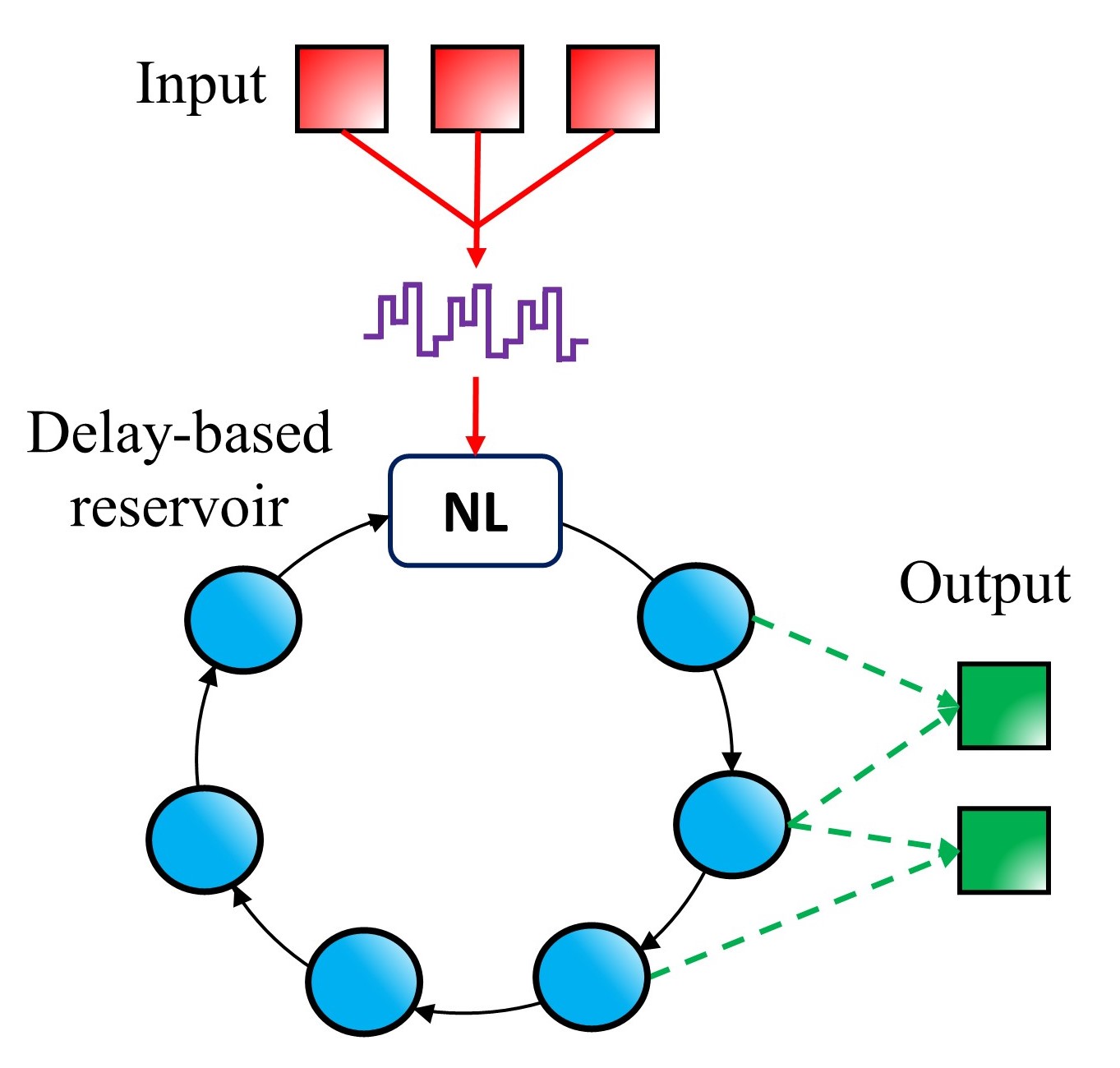}}
\caption{Architecture of a delay-based reservoir computer. The reservoir states are obtained by time-multiplexing the input signal using a periodic input mask (in purple) and a  Non-Linear (NL) node. The rest of the scheme is similar to a standard reservoir computer, where only the output weights are trained.}
\label{fig_RC_delay}
\end{figure}

In this work we use a delay-based reservoir as basic building block of our deep configuration. Delay-based RCs have been extensively studied in previous works, starting from \cite{appeltant2011information} and have been proven to yield excellent results albeit having a simple structure \cite{rodan2010minimum, brunner2013parallel, vinckier2015high}.
Its basic architecture is shown in Fig.\ \ref{fig_RC_delay}.

The idea is that a reservoir can be implemented using a single physical nonlinear node to generate temporally-separated reservoir nodes in a delay line; this is done by time-multiplexing the inputs with a time-periodic input mask signal. In this way, Eq.\ \eqref{eq_RC} becomes:
\begin{equation} \label{eq_T-M}
\begin{aligned}
x_0(n+1) &= \sin(\alpha x_{N-1}(n-1) + \beta M_0u(n+1)) \\
x_i(n+1) &= \sin(\alpha x_{i-1}(n) + \beta M_iu(n+1)) 
\end{aligned}
\end{equation}
with $i = 1,.., N-1$, where $\alpha$ is the feedback strength  and $\beta$ the input strength. The two hyperparameters $\alpha$ and $\beta$ are tuned to maximize the reservoir's performance, and depend on the physical system and the task under test. More considerations on their optimal values and how to choose them are given in Sec.\ \ref{sec_hp}.

Deep delay reservoirs can be built in analogy with Eq.\ \eqref{eq_DRC}: the output of one reservoir is used as input to the following reservoir, and all the reservoir states are used during train and the test phase.

\section{Experiment}
\label{sec_exp}

\begin{figure}[!b]
\centerline{\includegraphics[scale=0.5]{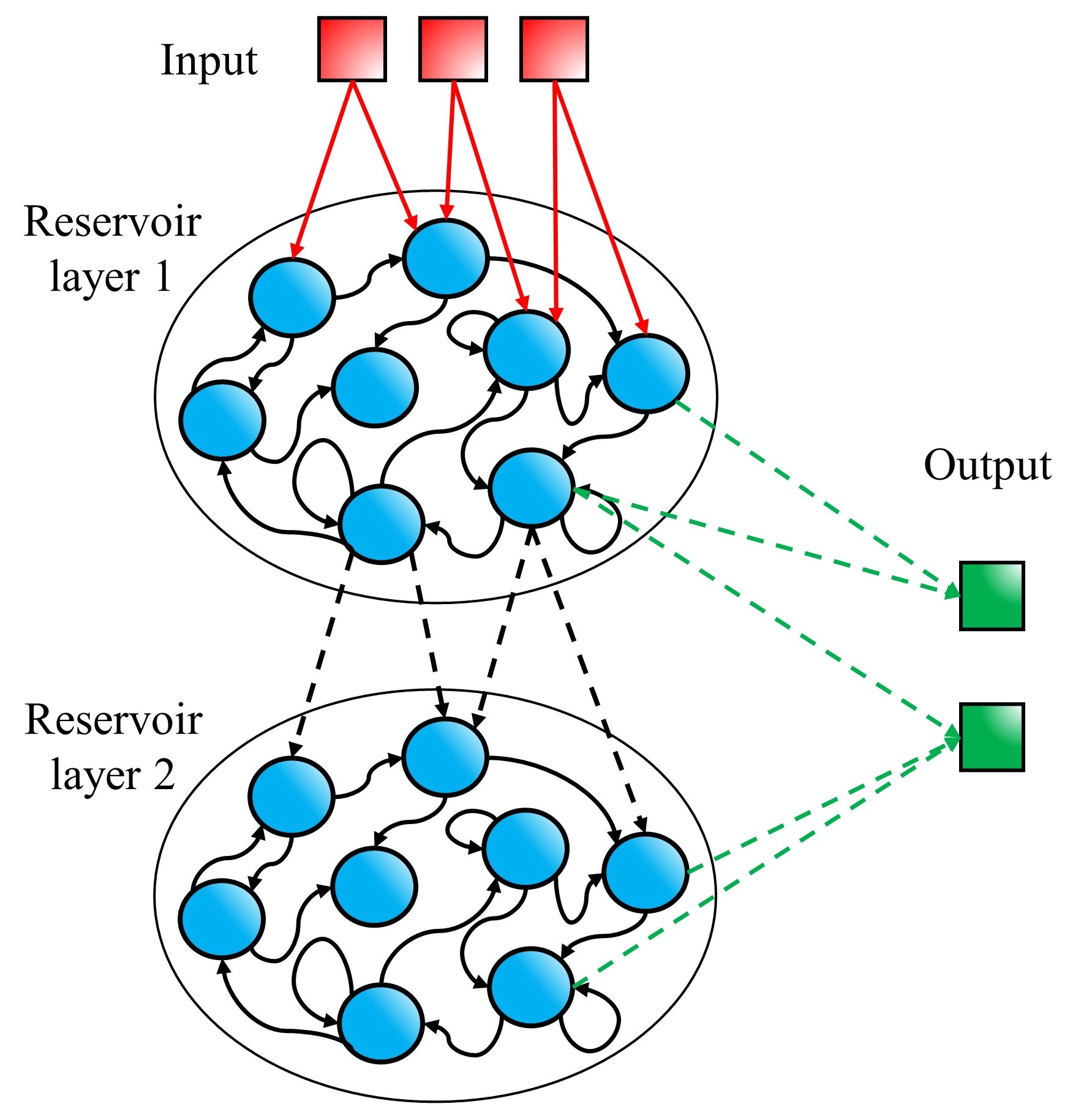}}
\caption{Architecture of a deep reservoir computer where the interlayer mask $\mathbf{
W}_2$ is optimized by means of an Evolutionary Algorithm. In this configuration both the interlayer and output weights are trained (dashed arrows). The input and internal reservoir connections are fixed (solid arrows). }
\label{fig_DRC_EA}
\end{figure}

\subsection{Evolutionary Algorithm applied to Deep Reservoir Computing}
\label{sec_EA}

One of the DRC configurations investigated in this work involves the optimization of the interconnections between different layers (cf.\ Fig.\ \ref{fig_DRC_EA}). To do so, we use a black box approach which belongs to the family of Evolutionary Algorithms, called Covariance Matrix Adaption-Evolution Strategy (CMA-ES, \cite{hansen2001completely}). 

The CMA-ES is based on the principles of natural evolution and adapts the covariance matrix of a multivariate Gaussian distribution to generate new candidate solutions for an optimization problem.
At each iteration, the CMA-ES algorithm generates a set of candidate solutions, called offsprings. The offsprings are then evaluated based on their fitness, and the algorithm updates the mean and covariance of the distribution based on the performance of the offspring. This update rule allows the algorithm to dynamically adapt to the structure of the problem and the geometry of the search space, leading to efficient exploration and exploitation of the search space. The algorithm continues this process until it reaches a stopping criterion, such as a maximum number of iterations or a desired level of fitness. 

In practical terms, for our experiments, the offsprings generated and evaluated by the CMA-ES are different interlayer masks ($\mathbf{W}_l$ in Eq.\ \eqref{eq_DRC}, with $l=2$). At every iteration of the algorithm, the CMA-ES generates a set of different offsprings, i.e.\ different interlayer matrixes $\mathbf{W}_2$. For every offspring, the deep reservoir is run and trained, and the performance is evaluated. Then, the best offspring of the current iteration becomes the base to generate the offsprings of the next iteration.

The immediate benefit of this approach is an improvement in the classification accuracy, at the cost of additional time to evaluate the performances of each offspring for every iteration. Because of this additional cost,  we only study the Evolutionary Algorithm approach for deep configurations with 2 layers. Experimental results, together with additional discussion, are reported in Sec.\ \ref{sec_results}.

\subsection{Optoelectronic setup}
\begin{figure}[!t]
\centerline{\includegraphics[scale=0.4]{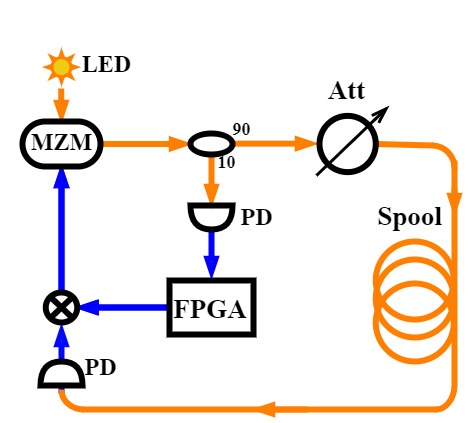}}
\caption{The experimental optoelectronic system used in this work. The optic fiber connections are in orange, and the electronic ones in blue.  LED: Superluminescent diode. MZM: Mach-Zender Intensity Modulator. PD: Photodetector. FPGA: Field Programmable Gate Array. Att: Optical Attenuator. Spool: 1.7km optic fiber spool. }
\label{fig_setup}
\end{figure}

The experimental setup used in this work is shown in Fig.\ \ref{fig_setup}, and is similar to the one used in \cite{antonik2016online} and more recently used in a deep configuration in \cite{nakajima2022physical}. It is an optoelectronic RC based on a Field Programmable Gate Array Board (FPGA) and a fiber-based photonic reservoir. 

The optical signal is generated by means of a superluminescent diode (\textit{Thorlabs SLD1550P-A40}) modulated by a Mach-Zehnder Intensity Modulator (MZM) (\textit{EOSPACE AX-2X2-0MSS-12}). The MZM modulates the light accordingly to the electrical signal received at its input, which is the sum of the masked input and the attenuated reservoir state at the previous timesteps; in this way Eq.\ \eqref{eq_T-M} is implemented, with the sinusoidal nonlinearity intrinsically implemented by the MZM characteristic. A 10\% fraction of the modulated light, which represents the reservoir state at present timestep, is sampled with a Photo-Detector (\textit{TTI TIA-525I}) and stored for the training and testing phase. The remaining 90\% of the light passes through an optical attenuator (\textit{JDS HA9}) and afterwards in a 1.7-km-long fiber spool. The time necessary for the light to travel in the spool is the time-delay of the system mentioned in Sec.\ \ref{sec_delay}, and corresponds to 7.94 $\mu$s. The output of the spool is the collected by a second Photo-Detector (\textit{TTI TIA-525I}), electrically summed with the masked input and used to drive the MZM.

The FPGA (\textit{Xilinx Virtex-7} on a \textit{Xilinx VC707} evaluation board) interfaces the experiment using a 14-bit ADC and 16-bit DAC, and communicates with a PC from which the user can control the experiment. The PC-FPGA connection is established via a high-speed PCIe custom-designed link. The deep reservoir is realized using the FPGA: the board samples the state matrix $\mathbf{X}^l$ of reservoir layer $l$, computes the matrix multiplication between $\mathbf{X}^l$ and $\mathbf{W}_l$, and applies the new inputs to the reservoir layer $l+1$. Refer to Sec.\ \ref{sec_results} for more detailed information about the processing speed of FPGA and how it affects the overall speed of the system.

\subsection{Asynchronization}
In this work we experimentally validate a new approach in the design of delay-based reservoir computers. The standard approach in delay-based RC is to select a clock operating frequency dependent on the time-delay, either by setting the clock-cycle and time-delay resonant \cite{appeltant2011information, ortin2017reservoir} or slightly detuned (or ``desynchronized'') \cite{paquot2012optoelectronic, antonik2018application}. As pointed out in a recent numerical study \cite{hulser2022role}, this restriction on the choice of the clock-cycle is not necessary: most clock cycles give good performance. The removal of this contraint simplifies the design of the system without affecting the performance.
In our experiments we select a clock-frequency of 205 MHz, which is approximately the frequency used for a desynchronized reservoir with a time-delay of 7.94 $\mu$s  and 200 internal nodes sampled 8 times each. If considering our reservoir as a "desynchronized" one, there would be 202.46 nodes stored in the delay.
This frequency is used in all the experiments, and no degradation in performance is observed.

\subsection{Hyperparameters and Bayesian optimization}
\label{sec_hp}

The choice of hyperparameters is a critical step in the design of a Reservoir Computer. Hyperparameters are values which define the operating condition of the system, thus influence its performance, but, differently from output weights, are not determined by the training process.
In the design of our DRC, we consider three hyperparameters:
\begin{itemize}
    \item $\alpha$, the feedback strength of Eq.\ \eqref{eq_T-M}, physically tuned by the optical attenuator;
    \item $\beta$, the input strength of Eq.\ \eqref{eq_T-M}, digitally tuned by the DAC of the FPGA; 
    \item $\lambda$, the regularization parameter of Eq.\ \eqref{eq_ridge}. 
\end{itemize}
The same values of $\alpha$ and $\beta$ are applied to all the reservoir layers.

The usual approach when searching for the optimal hyperparameters is the so-called grid scan, consisting of testing all the possible combinations of hyperparameters. This procedure, albeit simple, is highly time-consuming especially in the case of slow and big-scale experimental reservoirs \cite{antonik2019human, bueno2018reinforcement, antonik2021bayesian}. For this reason we use an optimized algorithm for the search, the so-called Bayesian optimization \cite{mockus1994application, brochu2010tutorial}. It has been already used successfully in previous numerical \cite{yperman2016bayesian} and experimental \cite{picco2023high} works on RC. 

\begin{figure*}[!t]
\centerline{\includegraphics[scale=0.55]{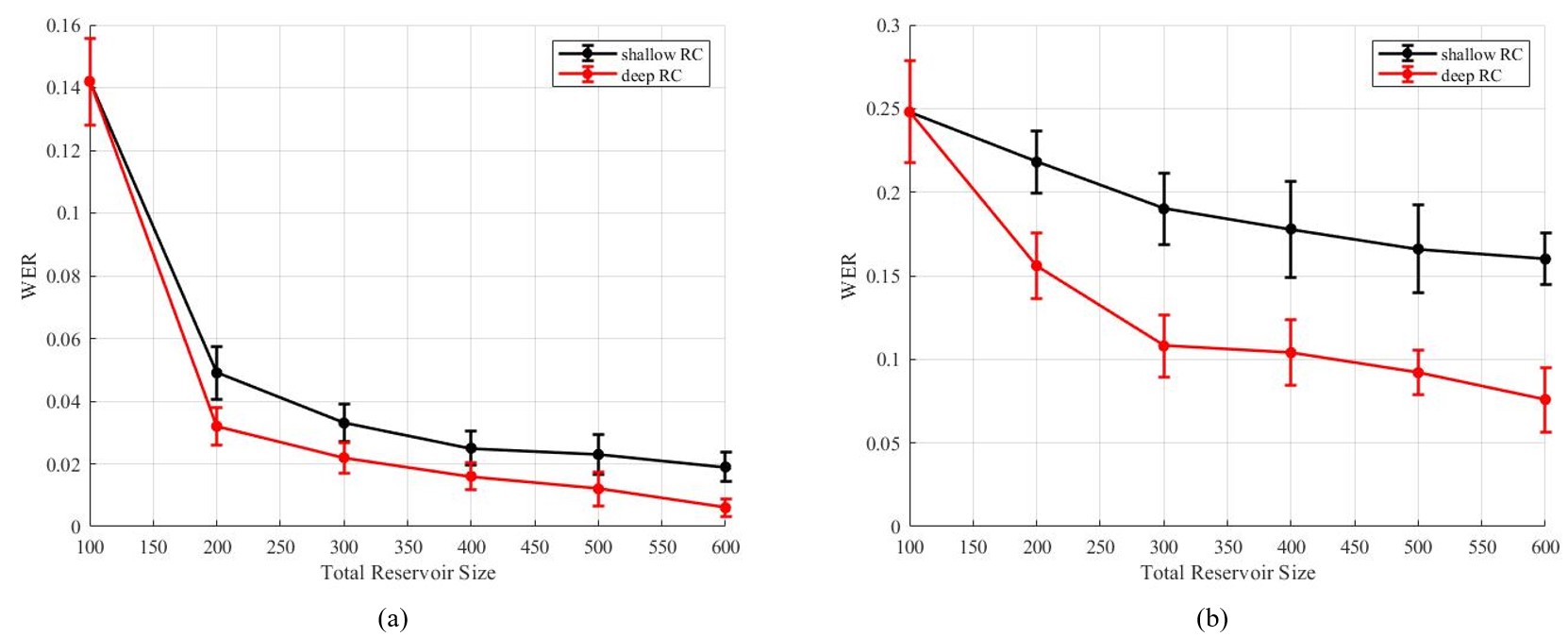}}
\caption{Experimental results obtained with our photonic reservoir computer on the spoken digits task. Results obtained on the original dataset are shown in figure (a), results obtained on the noisy dataset in figure (b). The classification accuracy, represented by the Word Error Rate (WER), is plotted as a function of the total number of trained parameters. For a shallow architecture, this number is equal to the reservoir nodes $N$; for a deep architecture with $L$ layers it amounts to $L\cdot 100$, since every reservoir layer contains 100 nodes. The black curves refer to the shallow architecture, the red curves to the deep architecture with random interlayer mask.}
\label{fig_plot_deep_vs_shallow}
\end{figure*}

Three components are needed to run the Bayesian optimization algorithm: a surrogate model, an acquisition function, and a strategy for selecting the next point to evaluate. 
The surrogate model is a probabilistic model that approximates the objective function based on the evaluations made so far. We use the Gaussian Process (GP) regression \cite{rasmussen2010gaussian} as surrogate model, while the objective function is the classification accuracy as a function of the hyperparameters. The model is updated as new evaluations are made and provides a probability distribution over the possible values of the objective function at unobserved points.
The acquisition function is used to balance the trade-off between exploration and exploitation of regions of the hyperparameter space. It determines the next point to evaluate by selecting the point that is most likely to improve the current best solution while also taking into account uncertainty in the surrogate model. In this work we use the Expected Improvement (EI) acquisition function.
The selection strategy is used to select the next point to evaluate based on the acquisition function. Here we use a sequential method, where hyperparameter points are selected one by one based on the results of previous evaluations. The process continues until a stopping criterion is met, such as reaching a maximum number of evaluations or a desired level of classification accuracy.

\subsection{Tasks}
\subsubsection{Spoken Digits Recognition}
\label{sec_SDR}

The Spoken Digit Recognition is a well known multi-class classification task used in previous works on RC \cite{martinenghi2012photonic, brunner2013parallel, verstraeten2007experimental, vinckier2015high}. 
The dataset is a subset of the NIST TI-46 corpus \cite{NIST}: it consists of 500 total utterances of the 10 (from 0 to 9) digits, repeated 10 times by 5 different subjects. 
We also use a second spoken digits dataset where a 3 dB Signal-To-Noise (SNR) ratio babble noise is present: in this way we can assess the capability of the system to work in the presence of noise.
The Lyon Passive Ear model \cite{lyon1982computational}, which models the biological response of the human auditory canal, is used to pre-process the audio signals. It transforms the time-domain utterances in frequency spectra with 86 channels. The frequency representations are then sent as input to the reservoir. 


\subsubsection{Japanese Vowels Classification}
The Japanese Vowels dataset \cite{kudo1999multidimensional} is widely used benchmark for time series analysis, and has been used in the past in the RC community \cite{paudel2020classification, yao2019intelligent, prater2017spatiotemporal}. It consists of a collection of 640 utterances of the Japanese vowel `ae', pronounced by 9 different male speakers. The task is thus to recognize the correct speaker for every utterance. The audio samples are pre-processed using the Mel-frequency cepstral coefficients (MFCCs), thus obtaining  a multivariate frequency representation for each utterance. Each sample consists then of 12 MFCC coefficients for every sample timestep, with sample lengths ranging from 7 to 29. The database is split in 270 train sequences and 370 test sequences. Similarly to the spoken digit dataset, these frequency representations are send to the reservoir as input signals. 

\section{Results and discussion}
\label{sec_results}

In this section we present our experimental results on the speech recognition task.
We investigate three different architectures:
\begin{itemize}
    \item a standard, or ``shallow'', reservoir computer with total number of internal nodes $N$ ranging from 100 to 600 for the spoken digit task, and from 50 to 300 for the Japanese vowels task;
    \item a deep reservoir computer with L reservoir layers of size $N=100$ each for the spoken digits task and $N=50$ for the Japanese vowels task, and random interlayer connections. The number of layers $L$ ranges from 1 to 6;
    \item a deep reservoir computer with 2 layers, where interlayer connections are optimized using the CMA-ES EA. 
\end{itemize}

The reason why different reservoir sizes are used for the two tasks is the following. The goal of this work is not to show the best achievable results that can be obtained using reservoir computing, or deep reservoir computing. Rather, it is to demonstrate the superiority of deep reservoir versus shallow reservoir.  The Japanese vowels task is considered an easier task to solve than the spoken digit task, and can be rather easily solved with accuracy close to perfection using a large reservoir.  For this reason, we preferred to investigate the Japan vowels task with a number of neurons suitable to investigate whether a different approach (i.e. deep reservoir computing) could enhance the performance. 

To compensate the limited size of the spoken digit recognition dataset (cf.\ Sec.\ \ref{sec_SDR}) and obtain some more reliable statistical validation, we use k-fold cross validation to train and test the system, with $k=10$. In k-fold cross validation, the dataset consisting of 500 utterances is divided in 10 equal subsets. The training is then repeated 10 times, using each time a different single subsets for the testing and the remaining 9 for the training; the classification accuracy is then averaged on the 10 training validations. 
In the case of the Japanese vowels task, we kept the same train/test division (270 train, 370 test) as previous works, for consistency. 
Then, instead of using k-fold cross validation, we run every experiment on 10 different input and interlayer masks, and average the results. 
The spoken digits task has 10 output classes: this means that 10 linear classifiers (cf.\ Sec.\ \ref{sec_RC}) are trained to output $+1$ for the predicted digit, or $-1$ otherwise, for every timestep. The output for a single utterance is then given by the most recurrent predicted output class during the duration of the utterance itself: this procedure is often referred to as winner-takes-all approach. The same procedure is used for the Japanese vowels dataset, but 9 classifiers are trained, since there are 9 output labels. 
For both tasks we use the Error Rate as figure of merit to evaluate the system's performance: it is simply the ratio between the digits/speakers predicted correctly and the total amount of digits/speakers. In the case of the spoken digit task, this is called the Word Error Rate (WER).
\begin{figure}[!b]
\centerline{\includegraphics[scale=0.35]{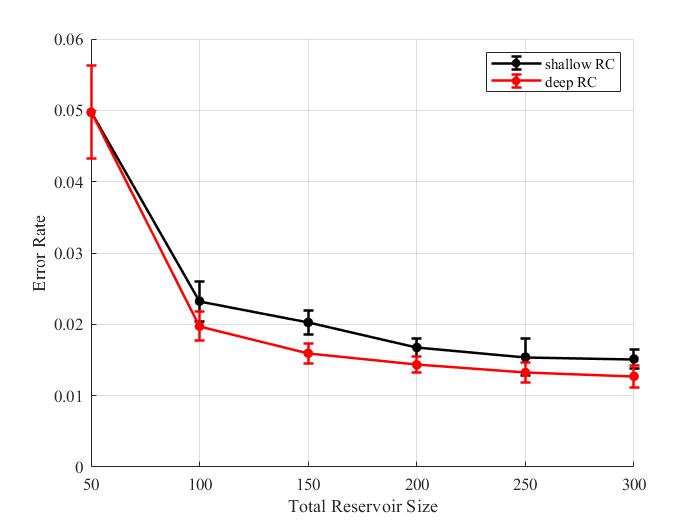}}
\caption{Experimental results obtained with our photonic reservoir computer on Japanese vowels classification. Similarly to Fig.\ \ref{fig_plot_deep_vs_shallow}, the classification accuracy, represented by the Error Rate, is plotted as a function of the total number of trained parameters. For a shallow architecture, this number is equal to the reservoir nodes $N$; for a deep architecture with $L$ layers it amounts to $L\cdot 50$, since every reservoir layer contains 50 nodes. The black curve refers to the shallow architecture, the red curve to the deep architecture with random interlayer mask.}
\label{fig_plot_deep_vs_shallow_jap}
\end{figure}

\begin{figure*}[b!]
\centerline{\includegraphics[scale=0.5]{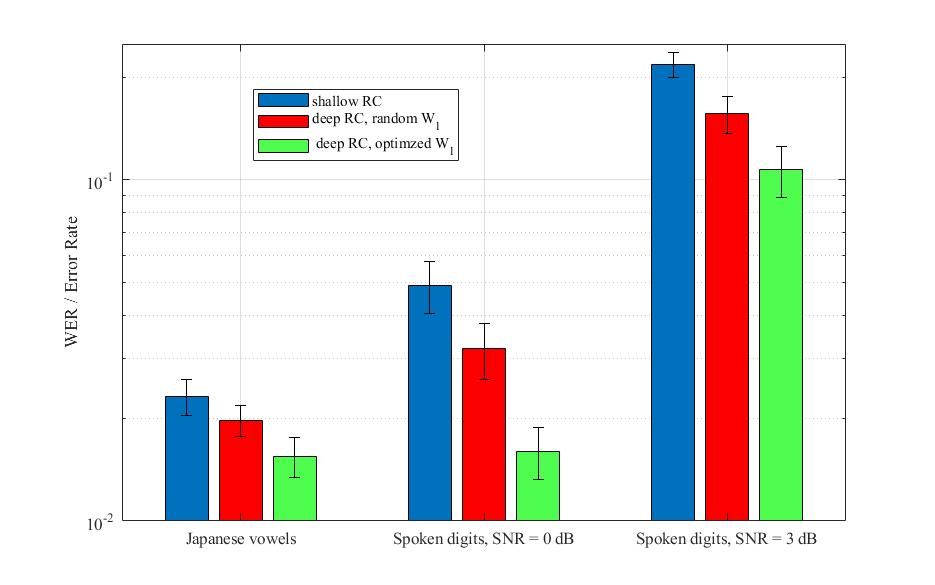}}
\caption{Experimental results on the spoken digits recognition task and the Japanese vowels classification task. This figures shows the results with three different architectures: shallow reservoir (blue), deep reservoir with random weights between the layers (red), and deep reservoir with interlayer weights optimized by means of an Evolutionary Algorithm (green). The performance are represented by the Error Rate (called WER for the spoken digits task). For the spoken digits task there are two different SNR values corresponding to the two datasets (noiseless and noisy) while only a noiseless dataset is available for the Japanese vowels recognition.
From the results is clear that not only a deep ``random'' reservoir performs better than the shallow network, as already presented in fig. \ref{fig_plot_deep_vs_shallow} and \ref{fig_plot_deep_vs_shallow_jap}, but the performance of the deep ``optimized'' reservoir outperforms the other two architectures.}
\label{fig_plot_GA_vs}
\end{figure*}

Fig.\ \ref{fig_plot_deep_vs_shallow} and Fig.\ \ref{fig_plot_deep_vs_shallow_jap} report our experimental results using a deep architecture with random interlayer connections, compared with a shallow architecture. Fig.\ \ref{fig_plot_deep_vs_shallow}(a) refers to the original spoken digits dataset, whereas Fig.\ \ref{fig_plot_deep_vs_shallow}(b) to the spoken digits dataset with noise. Fig.\ \ref{fig_plot_deep_vs_shallow_jap} refers to the Japanese vowels dataset. For every experimental point the hyperparameters are adjusted using Bayesian optimization (cf.\ Sec.\ \ref{sec_hp}) and the optimal result is shown. The results obtained with the shallow and deep reservoir are compared considering networks with the same number of internal variables, or nodes, so that the number of trained output weights are the same. In case of the shallow reservoir, the internal number of nodes is equivalent to $N$, whereas in our deep reservoir it is equal to the number of layers $L$ times the internal nodes of each layer, i.e.\ $L\cdot 100$ for the spoken digits task and $L\cdot 50$ for the Japanese vowels task. The Error Rate is plotted against the total number of internal nodes, and the trend is clear: first, the classification accuracy increase with the total number of nodes; second and most importantly, for the same number of nodes, the deep architecture always outperform the shallow one. 
Furthermore, the results show that the deep architecture has a better impact on the performance in the case of noisy input signal: the confidence intervals partially overlap in the case of the noiseless datasets (Fig.\ \ref{fig_plot_deep_vs_shallow}(a) and Fig.\ \ref{fig_plot_deep_vs_shallow_jap}), while they are well spaced when noise is present (Fig.\ \ref{fig_plot_deep_vs_shallow}(b)).
Gallicchio et al. \cite{gallicchio2018design} investigated the deep reservoir architecture on the same spoken digit tasks (without noise) used here, but with an artificial echo state network run on digital hardware. They present results very similar to Fig.  \ref{fig_plot_deep_vs_shallow}(a), showing the superiority of the deep architecture over the shallow one. In their DRC architecture, every reservoir’s layer has 50 nodes, while in our work 100 nodes are present in each layer. From the quantitative point of view, while we obtain a best WER of 0.006 using our maximum number layers (L = 6), they obtain a WER of approximately 0.004 using the same number of layers: a discrepancy most likely due to experimental noise in the optoelectronic setup, an issue not present in noiseless digital processors. 

Fig.\ \ref{fig_plot_GA_vs} shows our experimental results with the third architecture considered in this work, i.e.\ a deep reservoir computer composed of two reservoir layers where the interlayer connections are optimized using the CMA-ES algorithm (cf.\ Sec.\ \ref{sec_EA}), for both tasks under evaluation. As in the previous architecture, every reservoir has 100 nodes for the spoken digits task, and 50 nodes for the Japanese vowels task. 
In the same figures, these results are compared with (i) a shallow reservoir of the same total size as the deep reservoir (hence 200 nodes for spoken digits task and 100 nodes for Japanese vowels task),  and (ii) a deep reservoir made of two reservoir layers (again, with 100 or 50 nodes each, depending on the task), but with random interlayer connections. 
The classification error  is plotted for all the three datasets: spoken digits without noise, spoken digits with noise (SNR = 3 dB), and Japanese vowels. For both tasks, we can see that the deep ``random'' reservoir performs better than the shallow reservoir (as shown also in fig. \ref{fig_plot_deep_vs_shallow}); while the deep ``optimized'' reservoir clearly outperforms both of them. In our experiments, we observed that no more than 40 iterations are needed for the convergence of the CMA-ES algorithm. If the algorithm continues running after the optimum point is reached, the performance drops.
 Thus optimizing the connections between the layers is beneficial for the performance, at the cost of increased time needed for the convergence of the CMA-ES algorithm.

In terms of speed, the photonic reservoir can process the whole database in less than 3 ms, but it is limited by the slower electronics: the bottleneck is the 250 MHz bandwidth of the ADC.  Nonetheless, our optoelectronic reservoir computer can overcome the speed limitations of previous FPGA-based reservoir computers \cite{antonik2017brain} thanks to a combination of multiple factors. First, the PC-FPGA link is established with a custom designed PCIe interface, which allows to transfer data to the experiment at a rate of 2 gigasamples per second. Second, the Digital Signal Processing (DSP) slices on the FPGA are used for the computationally expensive matrix multiplications, such as the product between reservoir states $\mathbf{X}^l$ and interlayer mask $\mathbf{W}_l$. In this way the speed bottleneck due to the electronics is partially overcome. After training the system is able to process the audio recordings at high speed, inferring the classification of the whole datasets roughly twice faster than the duration of the audio signals, thus allowing to classify speech in real-time.

In conclusion, our results prove the effectiveness of using DRC for the recognition of human speech, both in an ideal and noisy environment. 



\section{Conclusion}
\label{sec_conclusion}

Deep Reservoir Computing (DRC) is emerging as a promising way to increase Reservoir Computing performance without resorting to fully-trained architectures. This not only reduces the power consumption related to training, but also allows the implementation of the algorithm into photonic substrates which could not support the tuning of each neuronal connection. We demonstrated the feasibility of optoelectronic DRC applied to human speech recognition.

We tested three architectures: standard RC, DRC with random interlayer weights and DRC with optimized interlayer weights. The ``optimized'' is the best in terms of performance, but additional iterations are needed for the optimization algorithm. The ``randomized'' DRC, although less accurate than the ``optimized'' one, still outperforms traditional RC. These results corroborate what has been experimentally measured also in \cite{lupo2023fully}. We remark that, contrary to the approach followed in \cite{nakajima2022physical}, our optimization scheme does not imply any back-propagation, neither requires a model of the substrate; moreover, we claim that, even when not optimized, DRC performs better than traditional RC.

We proposed design choices which make our optolectronic system robust, easy to design and efficiently adjustable in case of changes in external conditions and datasets. First, we introduced a simplified design for time-delay RC systems consisting of the asynchronization of clock-cycle and delay-time, based on the previous numerical work reported in \cite{hulser2022role}. Second, we employed a Bayesian optimization for the hyperparameters, which only requires a few tens of iterations. The system has been proved capable to performing real-time classification of audio signals, both in no-noise and in noisy environments.

Our results call for future studies on different aspects in the design of standard and deep RC, such as: (i) tuning the clock-cycle asynchronization as a hyperparameter, as suggested in \cite{hulser2022role}, (ii) tuning hyperparameters independently for each reservoir layers, (iii) optimizing the interlayer masks when more than two reservoirs are stacked in series, (iv) optimizing the input layer of the first reservoir, (v) using more efficient algorithms rather than the black-box CMA-ES approach for the selection of interlayer connections, and (vi) testing the DRC system with more complex time-series such as human action recognition on videos.

We thus hope with this work to open the way to the development of neuromorphic photonic hardware for high-speed and energy-efficient real world applications.

\section{Acknowledgments}
The authors acknowledge financial support from the H2020 Marie Skłodowska-Curie Actions (Project POSTDIGITAL Grant number 860360); and from the Fonds de la Recherche Scientifique - FNRS.

\bibliographystyle{IEEEtran}
\bibliography{IEEEabrv, DRC_journal}

\begin{thebibliography}{10}
\providecommand{\url}[1]{#1}
\csname url@samestyle\endcsname
\providecommand{\newblock}{\relax}
\providecommand{\bibinfo}[2]{#2}
\providecommand{\BIBentrySTDinterwordspacing}{\spaceskip=0pt\relax}
\providecommand{\BIBentryALTinterwordstretchfactor}{4}
\providecommand{\BIBentryALTinterwordspacing}{\spaceskip=\fontdimen2\font plus
\BIBentryALTinterwordstretchfactor\fontdimen3\font minus \fontdimen4\font\relax}
\providecommand{\BIBforeignlanguage}[2]{{%
\expandafter\ifx\csname l@#1\endcsname\relax
\typeout{** WARNING: IEEEtran.bst: No hyphenation pattern has been}%
\typeout{** loaded for the language `#1'. Using the pattern for}%
\typeout{** the default language instead.}%
\else
\language=\csname l@#1\endcsname
\fi
#2}}
\providecommand{\BIBdecl}{\relax}
\BIBdecl

\bibitem{reuther2020survey}
A.~Reuther, P.~Michaleas, M.~Jones, V.~Gadepally, S.~Samsi, and J.~Kepner, ``Survey of machine learning accelerators,'' in \emph{2020 IEEE High Performance Extreme Computing Conference (HPEC)}.\hskip 1em plus 0.5em minus 0.4em\relax IEEE, 2020, pp. 1--12.

\bibitem{mehonic2022brain}
A.~Mehonic and A.~J. Kenyon, ``Brain-inspired computing needs a master plan,'' \emph{Nature}, vol. 604, no. 7905, pp. 255--260, 2022.

\bibitem{jaeger2004harnessing}
H.~Jaeger and H.~Haas, ``Harnessing nonlinearity: Predicting chaotic systems and saving energy in wireless communication,'' \emph{Science}, vol. 304, no. 5667, pp. 78--80, 2004.

\bibitem{maass2002real}
W.~Maass, T.~Natschl{\"a}ger, and H.~Markram, ``Real-time computing without stable states: A new framework for neural computation based on perturbations,'' \emph{Neural Computation}, vol.~14, no.~11, pp. 2531--2560, 2002.

\bibitem{tanaka2019recent}
G.~Tanaka, T.~Yamane, J.~B. H{\'e}roux, R.~Nakane, N.~Kanazawa, S.~Takeda, H.~Numata, D.~Nakano, and A.~Hirose, ``Recent advances in physical reservoir computing: A review,'' \emph{Neural Networks}, vol. 115, pp. 100--123, 2019.

\bibitem{lukovsevivcius2009reservoir}
M.~Luko{\v{s}}evi{\v{c}}ius and H.~Jaeger, ``Reservoir computing approaches to recurrent neural network training,'' \emph{Computer Science Review}, vol.~3, no.~3, pp. 127--149, 2009.

\bibitem{jaeger2002adaptive}
H.~Jaeger, ``Adaptive nonlinear system identification with echo state networks,'' \emph{Advances in neural information processing systems}, vol.~15, 2002.

\bibitem{antonik2018using}
P.~Antonik, M.~Gulina, J.~Pauwels, and S.~Massar, ``Using a reservoir computer to learn chaotic attractors, with applications to chaos synchronization and cryptography,'' \emph{Physical Review E}, vol.~98, no.~1, p. 012215, 2018.

\bibitem{jalalvand2015real}
A.~Jalalvand, G.~Van~Wallendael, and R.~Van~de Walle, ``Real-time reservoir computing network-based systems for detection tasks on visual contents,'' in \emph{2015 7th International Conference on Computational Intelligence, Communication Systems and Networks}.\hskip 1em plus 0.5em minus 0.4em\relax IEEE, 2015, pp. 146--151.

\bibitem{paquot2012optoelectronic}
Y.~Paquot, F.~Duport, A.~Smerieri, J.~Dambre, B.~Schrauwen, M.~Haelterman, and S.~Massar, ``Optoelectronic reservoir computing,'' \emph{Scientific Reports}, vol.~2, no.~1, p. 287, 2012.

\bibitem{verstraeten2005isolated}
D.~Verstraeten, B.~Schrauwen, D.~Stroobandt, and J.~Van~Campenhout, ``Isolated word recognition with the liquid state machine: a case study,'' \emph{Information Processing Letters}, vol.~95, no.~6, pp. 521--528, 2005.

\bibitem{picco2023high}
E.~Picco, P.~Antonik, and S.~Massar, ``High speed human action recognition using a photonic reservoir computer,'' \emph{arXiv preprint arXiv:2305.15283}, 2023.

\bibitem{hermans2014optoelectronic}
M.~Hermans, J.~Dambre, and P.~Bienstman, ``Optoelectronic systems trained with backpropagation through time,'' \emph{IEEE Transactions on Neural Networks and Learning Systems}, vol.~26, no.~7, pp. 1545--1550, 2014.

\bibitem{hermans2015photonic}
M.~Hermans, M.~C. Soriano, J.~Dambre, P.~Bienstman, and I.~Fischer, ``Photonic delay systems as machine learning implementations,'' \emph{Journal of Machine Learning Research}, vol.~16, pp. 2081--2097, 2015.

\bibitem{hermans2015trainable}
M.~Hermans, M.~Burm, T.~Van~Vaerenbergh, J.~Dambre, and P.~Bienstman, ``Trainable hardware for dynamical computing using error backpropagation through physical media,'' \emph{Nature Communications}, vol.~6, no.~1, p. 6729, 2015.

\bibitem{hermans2016embodiment}
M.~Hermans, P.~Antonik, M.~Haelterman, and S.~Massar, ``Embodiment of learning in electro-optical signal processors,'' \emph{Physical Review Letters}, vol. 117, no.~12, p. 128301, 2016.

\bibitem{wright2022deep}
L.~G. Wright, T.~Onodera, M.~M. Stein, T.~Wang, D.~T. Schachter, Z.~Hu, and P.~L. McMahon, ``Deep physical neural networks trained with backpropagation,'' \emph{Nature}, vol. 601, no. 7894, pp. 549--555, 2022.

\bibitem{triefenbach2010phoneme}
F.~Triefenbach, A.~Jalalvand, B.~Schrauwen, and J.-P. Martens, ``Phoneme recognition with large hierarchical reservoirs,'' \emph{Advances in Neural Information Processing Systems}, vol.~23, 2010.

\bibitem{freiberger2019improving}
M.~Freiberger, S.~Sackesyn, C.~Ma, A.~Katumba, P.~Bienstman, and J.~Dambre, ``Improving time series recognition and prediction with networks and ensembles of passive photonic reservoirs,'' \emph{IEEE Journal of Selected Topics in Quantum Electronics}, vol.~26, no.~1, pp. 1--11, 2019.

\bibitem{gallicchio2017deep}
C.~Gallicchio, A.~Micheli, and L.~Pedrelli, ``Deep reservoir computing: A critical experimental analysis,'' \emph{Neurocomputing}, vol. 268, pp. 87--99, 2017.

\bibitem{ma2020deepr}
Q.~Ma, L.~Shen, and G.~W. Cottrell, ``{DeePr-ESN}: A deep projection-encoding echo-state network,'' \emph{Information Sciences}, vol. 511, pp. 152--171, 2020.

\bibitem{gallicchio2018design}
C.~Gallicchio, A.~Micheli, and L.~Pedrelli, ``Design of deep echo state networks,'' \emph{Neural Networks}, vol. 108, pp. 33--47, 2018.

\bibitem{gallicchio2016deep}
C.~Gallicchio and A.~Micheli, ``Deep reservoir computing: A critical analysis,'' in \emph{Proceedings of the 24th European Symposium on Artificial Neural Networks (ESANN)}, 2016, p. 497–502.

\bibitem{gallicchio2018short}
C.~Gallicchio, ``Short-term memory of deep {RNN},'' in \emph{Proceedings of the 26th European Symposium on Artificial Neural Networks (ESANN)}, 2018, pp. 25--27.

\bibitem{Na2023}
X.~Na, W.~Ren, M.~Liu, and M.~Han, ``Hierarchical echo state network with sparse learning: A method for multidimensional chaotic time series prediction,'' \emph{IEEE Transactions on Neural Networks and Learning Systems}, vol.~34, no.~11, pp. 9302--9313, 2023.

\bibitem{Pedrelli2022}
L.~Pedrelli and X.~Hinaut, ``Hierarchical-task reservoir for online semantic analysis from continuous speech,'' \emph{IEEE Transactions on Neural Networks and Learning Systems}, vol.~33, no.~6, pp. 2654--2663, 2022.

\bibitem{Chang2022}
H.-H. Chang, L.~Liu, and Y.~Yi, ``Deep echo state q-network (deqn) and its application in dynamic spectrum sharing for 5g and beyond,'' \emph{IEEE Transactions on Neural Networks and Learning Systems}, vol.~33, no.~3, pp. 929--939, 2022.

\bibitem{xu202111}
X.~Xu, M.~Tan, B.~Corcoran, J.~Wu, A.~Boes, T.~G. Nguyen, S.~T. Chu, B.~E. Little, D.~G. Hicks, R.~Morandotti \emph{et~al.}, ``11 {TOPS} photonic convolutional accelerator for optical neural networks,'' \emph{Nature}, vol. 589, no. 7840, pp. 44--51, 2021.

\bibitem{feldmann2021parallel}
J.~Feldmann, N.~Youngblood, M.~Karpov, H.~Gehring, X.~Li, M.~Stappers, M.~Le~Gallo, X.~Fu, A.~Lukashchuk, A.~S. Raja \emph{et~al.}, ``Parallel convolutional processing using an integrated photonic tensor core,'' \emph{Nature}, vol. 589, no. 7840, pp. 52--58, 2021.

\bibitem{liutkus2014imaging}
A.~Liutkus, D.~Martina, S.~Popoff, G.~Chardon, O.~Katz, G.~Lerosey, S.~Gigan, L.~Daudet, and I.~Carron, ``Imaging with nature: Compressive imaging using a multiply scattering medium,'' \emph{Scientific Reports}, vol.~4, no.~1, pp. 1--7, 2014.

\bibitem{saade2016random}
A.~Saade, F.~Caltagirone, I.~Carron, L.~Daudet, A.~Dr{\'e}meau, S.~Gigan, and F.~Krzakala, ``Random projections through multiple optical scattering: Approximating kernels at the speed of light,'' in \emph{2016 IEEE International Conference on Acoustics, Speech and Signal Processing (ICASSP)}.\hskip 1em plus 0.5em minus 0.4em\relax IEEE, 2016, pp. 6215--6219.

\bibitem{lugnan2020photonic}
A.~Lugnan, A.~Katumba, F.~Laporte, M.~Freiberger, S.~Sackesyn, C.~Ma, E.~Gooskens, J.~Dambre, and P.~Bienstman, ``Photonic neuromorphic information processing and reservoir computing,'' \emph{APL Photonics}, vol.~5, no.~2, p. 020901, 2020.

\bibitem{nakajima2022physical}
M.~Nakajima, K.~Inoue, K.~Tanaka, Y.~Kuniyoshi, T.~Hashimoto, and K.~Nakajima, ``Physical deep learning with biologically inspired training method: gradient-free approach for physical hardware,'' \emph{Nature Communications}, vol.~13, no.~1, p. 7847, 2022.

\bibitem{lupo2023fully}
A.~Lupo, E.~Picco, M.~Zajnulina, and S.~Massar, ``Deep photonic reservoir computer based on frequency multiplexing with fully analog connection between layers,'' \emph{Optica}, vol.~10, no.~11, pp. 1478--1485, Nov 2023.

\bibitem{appeltant2011information}
L.~Appeltant, M.~C. Soriano, G.~Van~der Sande, J.~Danckaert, S.~Massar, J.~Dambre \emph{et~al.}, ``Information processing using a single dynamical node as complex system,'' \emph{Nature Communications}, vol.~2, no.~1, p. 468, 2011.

\bibitem{larger2012photonic}
L.~Larger, M.~C. Soriano, D.~Brunner, L.~Appeltant, J.~M. Guti{\'e}rrez, L.~Pesquera, C.~R. Mirasso, and I.~Fischer, ``Photonic information processing beyond {Turing}: an optoelectronic implementation of reservoir computing,'' \emph{Optics Express}, vol.~20, no.~3, pp. 3241--3249, 2012.

\bibitem{Vettelschoss2022}
B.~Vettelschoss, A.~Röhm, and M.~C. Soriano, ``Information processing capacity of a single-node reservoir computer: An experimental evaluation,'' \emph{IEEE Transactions on Neural Networks and Learning Systems}, vol.~33, no.~6, pp. 2714--2725, 2022.

\bibitem{Köster2022}
F.~Köster, S.~Yanchuk, and K.~Lüdge, ``Master memory function for delay-based reservoir computers with single-variable dynamics,'' \emph{IEEE Transactions on Neural Networks and Learning Systems}, pp. 1--14, 2022.

\bibitem{Tang2023}
J.-Y. Tang, B.-D. Lin, Y.-W. Shen, R.-Q. Li, J.~Yu, X.~He, and C.~Wang, ``Asynchronous photonic time-delay reservoir computing,'' \emph{Opt. Express}, vol.~31, no.~2, pp. 2456--2466, Jan 2023.

\bibitem{Estebanez2023}
I.~Est\.{e}banez, A.~Argyris, and I.~Fischer, ``Experimental demonstration of bandwidth enhancement in photonic time delay reservoir computing,'' \emph{Opt. Lett.}, vol.~48, no.~9, pp. 2449--2452, May 2023.

\bibitem{Abdalla2023}
M.~Abdalla, C.~Zrounba, R.~Cardoso, P.~Jimenez, G.~Ren, A.~Boes, A.~Mitchell, A.~Bosio, I.~O'Connor, and F.~Pavanello, ``Minimum complexity integrated photonic architecture for delay-based reservoir computing,'' \emph{Opt. Express}, vol.~31, no.~7, pp. 11\,610--11\,623, Mar 2023.

\bibitem{hulser2022role}
T.~H{\"u}lser, F.~K{\"o}ster, L.~Jaurigue, and K.~L{\"u}dge, ``Role of delay-times in delay-based photonic reservoir computing,'' \emph{Optical Materials Express}, vol.~12, no.~3, pp. 1214--1231, 2022.

\bibitem{tikhonov1995numerical}
A.~N. Tikhonov, A.~Goncharsky, V.~V. Stepanov, and A.~G. Yagola, \emph{Numerical methods for the solution of ill-posed problems}.\hskip 1em plus 0.5em minus 0.4em\relax Springer Science \& Business Media, 1995, vol. 328.

\bibitem{rodan2010minimum}
A.~Rodan and P.~Tino, ``Minimum complexity echo state network,'' \emph{IEEE Transactions on Neural Networks}, vol.~22, no.~1, pp. 131--144, 2010.

\bibitem{brunner2013parallel}
D.~Brunner, M.~C. Soriano, C.~R. Mirasso, and I.~Fischer, ``Parallel photonic information processing at gigabyte per second data rates using transient states,'' \emph{Nature Communications}, vol.~4, no.~1, p. 1364, 2013.

\bibitem{vinckier2015high}
Q.~Vinckier, F.~Duport, A.~Smerieri, K.~Vandoorne, P.~Bienstman, M.~Haelterman, and S.~Massar, ``High-performance photonic reservoir computer based on a coherently driven passive cavity,'' \emph{Optica}, vol.~2, no.~5, pp. 438--446, 2015.

\bibitem{hansen2001completely}
N.~Hansen and A.~Ostermeier, ``Completely derandomized self-adaptation in evolution strategies,'' \emph{Evolutionary Computation}, vol.~9, no.~2, pp. 159--195, 2001.

\bibitem{antonik2016online}
P.~Antonik, F.~Duport, M.~Hermans, A.~Smerieri, M.~Haelterman, and S.~Massar, ``Online training of an opto-electronic reservoir computer applied to real-time channel equalization,'' \emph{IEEE Transactions on Neural Networks and Learning Systems}, vol.~28, no.~11, pp. 2686--2698, 2016.

\bibitem{ortin2017reservoir}
S.~Ort{\'\i}n and L.~Pesquera, ``Reservoir computing with an ensemble of time-delay reservoirs,'' \emph{Cognitive Computation}, vol.~9, no.~3, pp. 327--336, 2017.

\bibitem{antonik2018application}
P.~Antonik, \emph{Application of {FPGA} to Real-Time Machine Learning: Hardware Reservoir Computers and Software Image Processing}.\hskip 1em plus 0.5em minus 0.4em\relax Springer, 2018.

\bibitem{antonik2019human}
P.~Antonik, N.~Marsal, D.~Brunner, and D.~Rontani, ``Human action recognition with a large-scale brain-inspired photonic computer,'' \emph{Nature Machine Intelligence}, vol.~1, no.~11, pp. 530--537, 2019.

\bibitem{bueno2018reinforcement}
J.~Bueno, S.~Maktoobi, L.~Froehly, I.~Fischer, M.~Jacquot, L.~Larger, and D.~Brunner, ``Reinforcement learning in a large-scale photonic recurrent neural network,'' \emph{Optica}, vol.~5, no.~6, pp. 756--760, 2018.

\bibitem{antonik2021bayesian}
P.~Antonik, N.~Marsal, D.~Brunner, and D.~Rontani, ``Bayesian optimisation of large-scale photonic reservoir computers,'' \emph{Cognitive Computation}, pp. 1--9, 2021.

\bibitem{mockus1994application}
J.~Mockus, ``Application of {Bayesian} approach to numerical methods of global and stochastic optimization,'' \emph{Journal of Global Optimization}, vol.~4, pp. 347--365, 1994.

\bibitem{brochu2010tutorial}
E.~Brochu, V.~M. Cora, and N.~De~Freitas, ``A tutorial on {Bayesian} optimization of expensive cost functions, with application to active user modeling and hierarchical reinforcement learning,'' \emph{arXiv preprint arXiv:1012.2599}, 2010.

\bibitem{yperman2016bayesian}
J.~Yperman and T.~Becker, ``Bayesian optimization of hyper-parameters in reservoir computing,'' \emph{arXiv preprint arXiv:1611.05193}, 2016.

\bibitem{rasmussen2010gaussian}
C.~E. Rasmussen and H.~Nickisch, ``Gaussian processes for machine learning {(GPML)} toolbox,'' \emph{The Journal of Machine Learning Research}, vol.~11, pp. 3011--3015, 2010.

\bibitem{martinenghi2012photonic}
R.~Martinenghi, S.~Rybalko, M.~Jacquot, Y.~K. Chembo, and L.~Larger, ``Photonic nonlinear transient computing with multiple-delay wavelength dynamics,'' \emph{Physical Review Letters}, vol. 108, no.~24, p. 244101, 2012.

\bibitem{verstraeten2007experimental}
D.~Verstraeten, B.~Schrauwen, M.~d’Haene, and D.~Stroobandt, ``An experimental unification of reservoir computing methods,'' \emph{Neural Networks}, vol.~20, no.~3, pp. 391--403, 2007.

\bibitem{NIST}
Vol. Texas Instruments-Developed 46-Word Speaker-Dependent Isolated Word Corpus (TI46), September 1991, NIST Speech Disc 7-1.1 (1 disc).

\bibitem{lyon1982computational}
R.~Lyon, ``A computational model of filtering, detection, and compression in the cochlea,'' in \emph{ICASSP'82. IEEE International Conference on Acoustics, Speech, and Signal Processing}, vol.~7.\hskip 1em plus 0.5em minus 0.4em\relax IEEE, 1982, pp. 1282--1285.

\bibitem{kudo1999multidimensional}
M.~Kudo, J.~Toyama, and M.~Shimbo, ``Multidimensional curve classification using passing-through regions,'' \emph{Pattern Recognition Letters}, vol.~20, no. 11-13, pp. 1103--1111, 1999.

\bibitem{paudel2020classification}
U.~Paudel, M.~Luengo-Kovac, J.~Pilawa, T.~J. Shaw, and G.~C. Valley, ``Classification of time-domain waveforms using a speckle-based optical reservoir computer,'' \emph{Optics Express}, vol.~28, no.~2, pp. 1225--1237, 2020.

\bibitem{yao2019intelligent}
X.~Yao and Z.~Wang, ``An intelligent interconnected network with multiple reservoir computing,'' \emph{Applied Soft Computing}, vol.~78, pp. 286--295, 2019.

\bibitem{prater2017spatiotemporal}
A.~Prater, ``Spatiotemporal signal classification via principal components of reservoir states,'' \emph{Neural Networks}, vol.~91, pp. 66--75, 2017.

\bibitem{antonik2017brain}
P.~Antonik, M.~Haelterman, and S.~Massar, ``Brain-inspired photonic signal processor for generating periodic patterns and emulating chaotic systems,'' \emph{Physical Review Applied}, vol.~7, no.~5, p. 054014, 2017.

\end{thebibliography}

\end{document}